# Correcting Real-Word Spelling Errors: A New Hybrid Approach

Seyed MohammadSadegh Dashti(1), Amid Khatibi Bardsiri(2), Vahid Khatibi Bardsiri(3)

(1)(2)(3) Department of Computer Engineering, Kerman Branch, Islamic Azad University, Kerman, Iran
(1)dashti@iauk.ac.ir, (2)a.khatibi@sbriau.ac.ir, (3)kvahid2@live.utm.my

**Abstract**

Spelling correction is one of the main tasks in the field of Natural Language Processing. Contrary to common spelling errors, real-word errors cannot be detected by conventional spelling correction methods. The real-word correction model proposed by Mays, Damerau and Mercer showed a great performance in different evaluations. In this research, however, a new hybrid approach is proposed which relies on statistical and syntactic knowledge to detect and correct real-word errors. In this model, Constraint Grammar (CG) is used to discriminate among sets of correction candidates in the search space. Mays, Damerau and Mercer's trigram approach is manipulated to estimate the probability of syntactically well-formed correction candidates. The approach proposed here is tested on the Wall Street Journal corpus. The model can prove to be more practical than some other models, such as WordNet-based method of Hirst and Budanitsky and fixed windows size method of Wilcox-O'Hearn and Hirst.

## 1 Introduction

Proposing new methods for auto-correcting errors, particularly real-word ones, is a thorny task. Because of this challenge, many of the existing spelling correction methods are not sufficiently accurate. There are two primary undertakings in spelling correction: error detection and error correction. Errors found in general writings are roughly classified into two sets: non-word errors and real-word errors. Non-word errors may occur when the typist produces a misspelled word. As there is no correct spelling for this kind of errors, they may not be attested in the dictionary list. Therefore, spelling correction programs, including word processors, may effectively detect and subsequently correct errors. Moreover, real-word errors may occur when a user mistypes a correctly spelled word. Many spelling correction programs cannot recognize such errors, because they normally process words in isolation. This way, they can merely distinguish non-word errors which are mistyped and cannot be found in the dictionary. Real-word errors may also appear in cases where a word is mistyped because of its phonetic similarity to another word. Likewise, these errors may be committed when a word does not seem to match the context (e.g. sentence) in which it is used. In other cases, they may occur when the user tries to replace a non-word error, using the suggestion list in the text processing software. A real-word error takes place when a user mistakenly chooses the wrong word among the other alternatives recommended in the suggestion list (Wilcox-O'Hearn and Hirst, 2008). Even spelling correction programs may mistakenly generate a real-word error while attempting to adjust non-word errors, in cases where the "auto-correct" feature is enabled in the text processing setting (Hirst and Budanitsky, 2005; Dashti, 2017). Detailed investigation of real-word spelling corrections has been proposed by Pedler (2007) Hirst and Budanitsk (2005). Of course, Kukich (1992) has more extensively explored spelling correction. Dashti (2017) proposed a model which drew on a Probabilistic Context-Free Grammar and Wilcox-O'Hearn and Hirst's (2008) fixed length windows to correct multiple real-word errors in a sentence. This model showed significantly better performance in the evaluations. In this study, we propose a model for auto-correcting multiple real-word errors in a sentence. This model uses both statistical and syntactic knowledge to detect and correct real-word errors. The fixed-length windows model (Wilcox-O'Hearn and Hirst, 2008) and English Constraint Grammar (Voutilainen and Heikkilä, 1993) are used in the proposed approach. We clarify why our model identifies and auto-corrects real-word errors more efficiently, especially when the typist is not professional. We review the literature in this field (Wilcox-O'Hearn and Hirst, 2008; Hirst and Budanitsky, 2005) and uses the corpus of Wall Street Journal as a benchmark for the assessment. Using such a corpus would make it possible to compare the findings with those of similar works (Wilcox-O'Hearn and Hirst, 2008; Hirst and Budanitsky, 2005; Dashti, 2017). This paper is organized in the following order: in section 2, Mays and colleagues' (1991) model, which lays the foundation of our model, is briefly reviewed. In section 3, a thorough discussion on the existing important variations and the updates on Mays and colleagues' (1991) model are presented. Our proposed model is explicated in section 4. In the next section, evaluations and experimental results are reported. In section 6, an outline of the relevant studies in the literature is framed, and finally in section 7, concluding remarks about the discussion are provided.

## 2 Mays, Damerau and Mercer's model

In this section, Mays and colleagues' (1991) model is reviewed and some of its important shortcomings are discussed.

### 2.1 Overview of the model

Mays and colleagues' (1991) approach represented a sample of the noisy-channel model. It was designed to correct the sentence $S$, which went through a noisy channel (e.g. a typist) that could create a few errors by mistake, in the correct sentence $S'$. Mays et al. (1991) considered α to be a parameter which represented the probability of a word's correct spelling. The remaining fraction (1-α) showed the probability that the word was mistyped as a real-word error, showing that this probability was equally distributed among all the related correction candidates. Therefore, the probability that the intended $w$ was typed as $y$ was formulated as follows:

(1)
$$P(w|y) \begin{cases} \alpha & \text{if } y = w \\ (1-\alpha)/|S_c(w)| & \text{if } y \in S_c(w) \end{cases}$$

In equation 1, $S_c(w)$ is the spelling variations of $w$ generated by the *ispell* software (Kuenning et al., 2004). Afterwards, equation 2 was used to estimate the probability of every single related trigram, and then to replace the most probable one with the original trigram:

(2)
$$P(S') = \prod_{i=1}^{n} P(w_i | w_{i-1} w_{i-2})$$

In this configuration, $S'$, which is one of the current sentences in the search space $C(S')$, maximizes the probability of $P(S'|S) = P(S').P(S|S')$.

### 2.2 Disadvantages of Mays, Damerau and Mercer's method

One of the disadvantages of this model is the large size of the trigram model, which is an essential element to produce an effective performance. The other disadvantage is that the model tries to correct grammatical errors which can be detected and corrected through grammar checkers only. According to Wilcox-O'Hearn and Hirst (2008), there is an undesirable shortcoming in this model: because each member of the search space $C(S)$ is a sentence with one word changed in it, the method would be able to correct only one real-word error in each sentence. This process could be more challenging if the α value showed a low degree, especially in cases where the source of the error (e.g. typist) might produce many spelling errors. Consider the phrase "the two of them" which the user intended to type, although by mistake s/he typed "thew to of then". In this sequence of words, $w_2$= *thew*, $w_2$= *to*, $w_3$= *of*, and $w_4$= *then*. According to Mays et al., (1991), in such a situation the model would create a search space of all the related word-sequences, each containing only a single spelling variation of the original phrase words. In short, the search space would include these word sequences:

$$C(S') = \{(S_c(w_1)\ w_2\ w_3\ w_4).(w_1\ S_c(w_2)\ w_3\ w_4). \\ (w_1\ w_2\ S_c(w_3)\ w_4).(w_1\ w_2\ w_3\ S_c(w_4)).\}$$

In addition, the model would try to estimate the probability of each sequence of words in the current search space, and replace the one with the highest probability value with the original phrase. However, as there is no word-sequence with more than one spelling variation through the search space, no exact replacement for the original phrase could be found.

## 3 Variations and improvements on the model

Wilcox-O'Hearn and Hirst (2008) introduced another assessment of the model, testing the model on the corpus of the WallStreet Journal. Although the model performed quite well in contrast with Hirst and Budanitsky's (2008) method, there was still some potential for improvement. In the present study, two important works are discussed that can help improve the model of Mays et al. (1991): Wilcox-O'Hearn and Hirst (2008) and Dashti (2017).

### 3.1. Fixed sized windows method suggested by Wilcox-O'Hearn and Hirst (2008)

The model of Mays et al. (1991) can typically make only one correction in each sentence. Wilcox-O'Hearn and Hirst (2008), considering it to be a NP hard problem, incorporated sentences with more than one correction in the search space. As they observed, such a capability would be helpful only when the typist was very careless, or in the case where the α value was very low. To solve this issue, Wilcox-O'Hearn and Hirst (2008) suggested a strategy. They tried to select all of the sentences which showed a higher probability value than the original sentence $S$ and to use a combination of them rather than a single sentence from the search space $C(S')$.

### 3.2. Using windows of fixed length

In the model of Mays et al. (1991), the sentences are usually used as variable-length units in order to be more optimized. Wilcox-O'Hearn and Hirst (2008) suggested a variation of the model which improved itself through windows of fixed length. In this variation at the first stage, the boundaries of the sentence are taken into consideration as *BoS* (Beginning-of-Sentence) and *EoS* (End-of-Sentence). Then a window with the fixed length $d+4$ ($d$ is the range of words) is used to accommodate the trigrams which are overlapped with the words in the current range. As a consequence, the smallest windows size will be 5, which incorporates three trigrams in estimating the probability of all the spelling variations of the middle word in the range. Thereafter, the method moves $d$ words to one side and checks the rest of the words in the sentence. Thus allowing multiple corrections in the sentence would be possible. If we consider the length of the sentence (including sentence markers, *BoS* and *EoS*), $l-d+1$ iteration(s) will be required to check the whole sentence.

### 3.3. Weaknesses of Wilcox-O'Hearn and Hirst's (2008) model

In recent years, unlike 1980s and mid-1990s, PCs and electronic devices are tremendously accessible to the majority of people around the world (Dashti, 2017). Most people use different types of electronic devices to type their content. As Dashti (2017) observes, every individual enjoys his\her own specific levels of attitude, accuracy and typing speed. From a practical perspective, various programs should be devised for different types of users, whether they are professional typists or beginners who have started typing on their handheld devices. As claimed by Wilcox-O'Hearn and Hirst (2008), multiple corrections in a sentence would not be exceptionally essential, although in the real world their idea is not true. Suppose α=.9, which is a typical value for this parameter. This value, implies that in every ten words typed by the user, one is not correct on average. In the real world, a user may normally mistype two words. The typos might occur one after the other or with a distance of a few words (Dashti, 2017). As the model of Mays et al. (1991) suggests, it only applies to one error in a sentence, although in reality it may fail to correct multiple real-word errors. As explained above, Wilcox-O'Hearn and Hirst (2008) proposed a variation which used windows of fixed length. One advantage of windows of fixed length is that they allow for multiple corrections in a sentence. Nevertheless, multiple corrections cannot be made in a particular window (Dashti, 2017). For example, consider the following sentence from the WallStreet Journal corpus:

*...test [tests] comparing its potpourri covert → convert [cover] with the traditional...*

This is an example of false-positive correction in a fixed-size window, in which "test" and "covert" are real-word mistakes. As can be seen, Wilcox-O'Hearn and Hirst's (2008) fixed-window failed to identify the first real-word error. Moreover, the model was unsuccessful in correcting the second error, because the method merely detects and corrects one error in each window.

### 3.4. Correcting multiple real-word errors in a window

Dashti (2017) made a thorough overview of the drawbacks in models proposed by Mays et al. (1991) and Wilcox-O'Hearn and Hirst (2008). He described why the variation formulated by Wilcox-O'Hearn and Hirst (2008) failed to correct multiple real-word errors in everyday real world conditions where α shows lower values. Dashti's (2017) proposed approach drew on Wilcox-O'Hearn and Hirst's (2008) fixed window size approach as a basis to create a search space including all the windows that were accommodated in a sentence. For every word $w_n$ in the current window, a group of spelling variations $S_c(w_n)$ might be considered. $CS_i$ is only a combined sequence of the words, in a current search space. $C(S'')$ represents the search space of the current window, incorporating all possible combinations of the current words and their related spelling variation sets. It should be noted that $CS_i \in C(S'')$.

Dashti (2017) manipulated a probabilistic context-free grammar (PCFG) (Klein and Manning, 2003) to discriminate between items in the search space. Any word sequence which had lower parse probability than the original word-sequence was removed from the search space. When the PCFG was applied, only syntactically well-formed word-sequences remained. Following that, the trigram approach proposed by Mays et al. (1991) was relied on to estimate the probability of the syntactically well-formed word-sequences. Next, word-sequences of all windows were combined with respect to their order of appearance, and the PCFG was applied to all the combinations. The combinations with higher parse probabilities than those of the original sentence were regarded as the final correction candidates. Consequently, equation 3 was used to estimate the probabilities of best correction candidates.

(3)
$$P(Combination) = \prod_{i=1}^{n}\prod_{j=1}^{m} P(CS_{ij})$$

where $CS_{ij}$ is the *i*-th syntactically well-formed word-sequence in the *j*-th window.

### 3.5. Advantages and weaknesses of Dashti's (2017) model

Dashti's (2017) model performed significantly better in evaluations, compared with the performances of Wilcox-O'Hearn and Hirst's (2008) and Hirst and Budanitsky's (2005) models. The difference of accuracy between Dashti's model and the others was even more evident in cases where α had lower values. Although the results were satisfying and the task of correcting multiple real-word errors was successful, the model was rather slow in processing everyday functions. This shortcoming appears to be a consequence of using Klein and Manning's (2003) PCFG model, which works rather slowly. Dashti (2017) stated that performance may be further improved by using high-speed unlexicalized PCFGs, such as the one proposed by Petrov et al. (2006). Dashti also pointed out that CG may be used to further improve the task of real-word error correction. In present research, CG is used to create a model for detecting and correcting multiple real-word errors in a window.

## 4 Methodology of the proposed hybrid approach

The proposed hybrid approach manipulates both statistical and syntactic knowledge to detect and correct real-word errors. In this extended model, Dashti's (2017) approach is relied on to generate the search space $C(S'')$ for each window. In the proposed model, CG is employed to discriminate among a set of correction candidates in the final search space. Then, Mays and colleagues'

(1991 trigram approach is used to estimate the probabilities of syntactically well-formed correction candidates. Section 4.1 details the CG and the modules which are used in our model. In section 4.2 we give a thorough discussion of how the proposed model works.

## 4.1. Constraint Grammar

In this section, we explain the English Constraint Grammar Parser (ENGCG) Voutilainen, and Heikkilä (1993). The first version of the ENGCG was developed by Voutilainen, Heikkilä and Anttila Voutilainen et al. (1992), based on the CG theory of Karlsson (Karlsson 1990, 1995). In following sub-section, the components of CG, which are used in the proposed model, are further explored.

### 4.1.1. Preprocessing

The initial stage in parsing is called *preprocessing*, which includes a variety of tasks such as: (a) the identification of sentence boundaries; (b) identification of punctuation marks; (c) identification of certain compounds, multiword prepositions, and other colloquial language structures; and (c) the normalization of certain orthographical patterns. The preprocessing component is developed as a set of approximately 7000 rewrite rules, most of which are fixed syntagms in the "BETA" programming language" Brodda (1990).

### 4.1.2. Morphological Analyzer

The main component in the morphological analyzer is *morphosyntactic* lexicon, which was designed based on Koskenniemi's well-known Two-Level model Koskenniemi (1983). Presently, the English lexicon ENGTWOL Karlsson et al. (1995), which contains approximately 84,000 lexical entries, represents the main vocabulary of modern English. ENGTWOL includes all inflected and central drivational English word-forms. Prefixes and endings are represented in separate 'minilexicons' which might be accessed from the 'stem' lexicon. Moreover, ENGTWOL manipulates a feature system which is mainly based on Quirk et al. (1985), including 139 morphosyntactic tags. Some of tags include parts of speech (*pos*) and others involve numbers, cases, moods, and so on. Moreover, they might also include essentially syntactic properties Heikkilä (1995). The ENGTWOL lexicon is performed as a two-level program called *twol*. Depending on the type of the text, the ENGTWOL analyzer identifies up to 99% of all running-text word-form tokens. For each token, at least one or more morphological analyses are conducted.

### 4.1.3. Morphological Disambiguation

For about 35–50% of all words in the input sentences, the morphological analyzer generates several alternative analyses. However, typically only one analysis is the proper match for the context. The morphological disambiguator recognizes the correct alternative by removing as many contextually illegitimate alternatives as possible. Optimally, the morphological disambiguator identifies unambiguous and correctly tagged sentences. However, this goal, from a practical perspective, is extremely difficult to achieve in the analysis of unrestricted texts. ENGCG relinquishes only those alternatives which represent a very small risk of error. Because the few most complicated cases are left pending, the output of the tagger is somewhat ambiguous. Most morphological or part-of-speech (*pos*) disambiguators (Church, 1988; Leech *et al.*, 1994; de Marcken, 1990) draw on co-occurrence-based and lexical statistics, which are usually derived from manually tagged corpora. In contrast. ENGCG only manipulates hand-written rules of language, or constraints which impose restrictions on the linear order of words and tags. Generally these types of constraints are very partial expressions of syntactic statements Voutilainen (1994). Generally they appear in the form: "remove reading Z if all context conditions are satisfied; else leave Z unchanged." The contextual conditions usually involve the fixed position of the words (e.g. "the second word to the left contains the tag Z") or the unbounded context within the sentence (e.g. "to the right, there is no X").

Some of the constraints are explained briefly to clarify the general idea:

1) Remove all finite verb readings if the preceding word is an unambiguous determiner.

2) Remove all subjunctive readings unless the left-hand context contains *that* or *lest* as a subordinating conjunction.

3) Remove all finite verb readings if the preceding word is "*to*".

The present grammar mainly includes two sections, the *grammar-based* section and the optionally applicable *heuristic* section. The first section contains about 1,150 constraints, making approximately 93-97% of all words unambiguous, out of which at least 99.7% showed correct morphological analyses. In the morphological disambiguation process, through the 200-odd heuristic constraints, 96–98% of all words were rendered unambiguous, however at this stage only about 99.5% showed correct morphological analyses.

### 4.2 The Proposed Hybrid Method in Use

Unlike the approaches suggested by Mays et al. (1991) and Wilcox-O'Hearn and Hirst (2008), the proposed design not only uses statistical knowledge but also manipulates syntactic knowledge to improve the process of detecting and correcting real-word errors. Dashti's (2017) approach (see section 3.4 above) was used to generate the search space $C(S)''$ for each window. Then, all $C(S)''$'s were combined according to windows' order to yield the final search space $C(S)$. $CS_k$ is a member of $C(S)$ ($CS_k \in C(S'')$); and it has the same length as the original sentence, according to expectations. Next, CG was applied to the $C(S)$ and syntactically ill-formed $CS_k$s were removed from $C(S)$. Finally Mays and colleagues' (1991) trigram approach was applied to the remaining correction candidates incorporated in $C(S)$, and a $CS_k$ with maximum value probability was selected as the best correction candidate. The whole process is briefly described in the following steps:

*Step 1.* Generate the search space $C(S)''$ for each window.
*Step 2.* Combine search spaces $C(S)''$ according to the windows' order, and create the final search space $C(S)$.
*Step 3.* Use ENGTWOL morphological analyzer to recognize all the word-form tokens of all $CS_k$s in $C(S)$ and make all possible morphological analyses.
*Step 4.* Employ the morphological disambiguator to identify the unambiguous and correctly tagged candidates ($CS_k$) by removing as many contextually illegitimate alternatives as possible.
*Step 5.* Use Mays and colleagues' (1991) trigram approach to identify the most probable correction candidate with the highest probability.

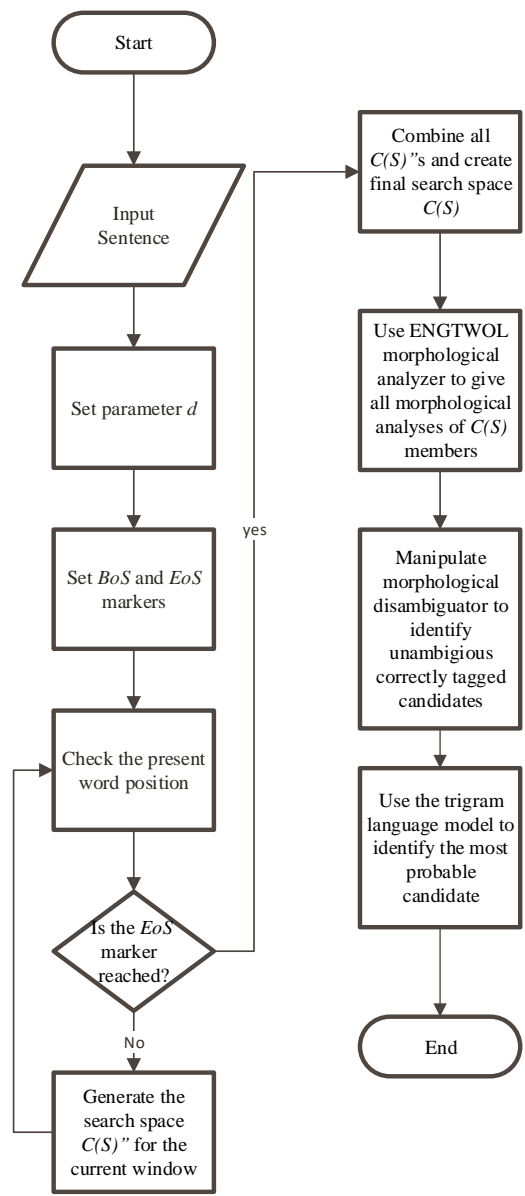

Fig. 1 Flowchart diagram of the model
Figure 1 dissertates the flowchart of the model constructed.

# 5 Evaluations

In this section, some of the most important disadvantages of Mays and colleagues' (1991) model are pointed out. Then, the proposed approach in this study is evaluated through measures of precision, recall, F-measure, and performance.

## 5.1 Reviewing the Evaluation in Mays et al. (1991)

Mays and colleagues' (1991) model was assessed using trigrams, the vocabulary of which included 20,000 words. Bahl et al. (1983) and Mays et al. (1991) implied that the probabilities were yielded from IBM Laser Patent Corpus, while the vocabulary of this corpus includes only 12,000 words. Another disadvantage of the model of Mays et al. (1991) is that the test data used were composed of 100 sentences, which were extracted from Transcripts of the Canadian Parliament and AP newswire (50 from each), although these sources are not presently available. In the assessment made by Mays et al. (1991), for a sentence $S'$ there was a search space of 86 sentences $S$, each containing only one error. In their assessment, no detail was shared about the length of the sentences. Additionally, the evaluation was not provided in terms of precision, recall and f-measure. The test and training data were not available to estimate the accuracy and performance of the measures. Due to these disadvantages, the assessment provided by Mays et al. (1991) is neither acceptable nor it might be compared with other models, namely those of Hirst and Budanitsky (2005) and Golding and Roth (1999). In the following section, the assessment of the hybrid mode is reported.

## 5.2 Assessment of Hybrid Approach

Wilcox-O'Hearn and Hirst (2008) did not just introduce a new variation that allowed for few corrections in a sentence by using windows of fixed length, but they proposed a new assessment of Mays and colleagues' (1991) method. Wilcox-O'Hearn and Hirst (2008) and Dashti, (2017) followed Hirst and Budanitsky's (2005) work and used 1987–89 Wall Street Journal corpus in their assessments. We also chose the similar corpus as Wilcox-O'Hearn and Hirst (2008) and Dashti (2017) did; the corpus incorporates about 30 million words, in which all the identifiers and headings are removed. Moreover, we assumed there was no error in the text, as it was. The test set was composed of 548 articles, each ranging from 115 to roughly 2,723 tokens, with the sum of 330,000 words. Following Wilcox-O'Hearn and Hirst (2008) and Dashti (2017), Cambridge Statistical Language Modeling Toolkit (Clarkson and Rosenfeld, 1997) was used to develop the trigram model. Wilcox-O'Hearn and Hirst (2008) evaluated their model using a 20,000 and a 62,000-word corpus. Evaluation results demonstrated that the model, which included 62,000 words, was by far more successful in detecting and correcting real-word errors (Dashti, 2017). Considering this fact, Dashti, (2017) used a 62,000 word data set in his evaluation. The vocabulary in this study, too, included 62,000 most frequently used words in the corpus, after the test data were omitted. Following Wilcox-O'Hearn and Hirst (2008), from the 548 saved articles we gathered, we randomly chose 33,508 sentences. After that, we made two separate test sets, each composed of 16,754 sentences detailed as follows:

**S62000:** Any word, in the 62,000 most frequent words in the corpus, can be replaced with one of its spelling variations from the same vocabulary.

**MALP:** Any word whose base form is a noun in the database of WordNet (without considering syntactic analysis) can be replaced with any of its spelling variations. This would replicate the malapropism data in Hirst and Budanitsky's (2005) model.

We chose four distinctive values of α in order to test the model: the values ranged from .9, which represented an unskilled typist, to .999, which replicated a precise typist. Depending on the current α value, different words may be replaced with their spelling variations. For example, for α =.95, a value already used by Hirst and Budanitsky (2005), about one in twenty words may be replaced by one of its spelling variations. A spelling variation is described as any word with the maximum of edit-distance of 1 from the original word, which can be an insertion, deletion, substitution or the transposition of two characters that leads to a real-word error. We evaluated the results, manipulating three measures: precision, recall and F-measure

(which was the harmonic mean of recall and precision). In order to calculate F-measure, we used equation 4:

(4)
$$F - measure = 2 * \frac{P * R}{P + R}$$

Three measures are shown for both the correction and deletion of errors in each table. We re-implemented the approaches suggested by Hirst and Budanitsky (2005), Wilcox-O'Hearn and Hirst (2008), and Dashti (2017), comparing them with our proposed model. In doing so, the same test and training data were employed as explained earlier. Table 1 reports the comparison of the results (*d*=1). The performance of the model is significant in both S62000 and MALP test sets. As expected, in lower values of α, particularly when α =.9, F-measure, recall and precision values for both correction and detection, showed a considerable increase in contrast with the findings of Wilcox-O'Hearn and Hirst (2008) and even of Dashti (2017).

|   | Detection | | | Correction | | |
|---|---|---|---|---|---|---|
| α | P | R | F | P | R | F |
| **Hirst and Budanitsky (2005)** | | | | | | |
| Test set **MALP**: | | | | | | |
| .95 | .231 | .312 | .265 | .212 | .289 | .244 |
| **Wilcox-O'Hearn and Hirst, (2008)** | | | | | | |
| Test set **S62000,** *d*=1: | | | | | | |
| .9 | .273 | .859 | .414 | .266 | .832 | .403 |
| .99 | .502 | .779 | .610 | .496 | .764 | .601 |
| .995 | .574 | .752 | .651 | .568 | .739 | .642 |
| .999 | .733 | .679 | .704 | .733 | .674 | .702 |
| Test set **MALP,** *d*=1: | | | | | | |
| .9 | .176 | .610 | .273 | .169 | .585 | .262 |
| .99 | .367 | .547 | .439 | .364 | .529 | .431 |
| .995 | .433 | .513 | .469 | .428 | .497 | .459 |
| .999 | .610 | .448 | .516 | .604 | .438 | .507 |
| **Multiple Corrections per Window** | | | | | | |
| Test set **S62000,** *d*=1: | | | | | | |
| .9 | .400 | .898 | .553 | .386 | .867 | .534 |
| .99 | .519 | .804 | .630 | .512 | .778 | .617 |
| .995 | .582 | .766 | .661 | .577 | .762 | .656 |
| .999 | .744 | .684 | .712 | .738 | .673 | .704 |
| Test set **MALP,** *d*=1: | | | | | | |
| .9 | .291 | .624 | .396 | .283 | .598 | .384 |
| .99 | .382 | .549 | .450 | .376 | .530 | .439 |
| .995 | .448 | .521 | .481 | .442 | .506 | .471 |
| .999 | .613 | .454 | .521 | .611 | .444 | .514 |
| **Hybrid Approach** | | | | | | |
| Test set **S62000,** *d*=1: | | | | | | |
| .9 | .774 | .967 | .859 | .769 | .947 | .848 |
| .99 | .805 | .948 | .870 | .798 | .939 | .862 |
| .995 | .817 | .937 | .872 | .810 | .931 | .866 |
| .999 | .849 | .925 | .885 | .846 | .908 | .875 |
| Test set **MALP,** *d*=1: | | | | | | |
| .9 | .775 | .956 | .856 | .771 | .938 | .846 |
| .99 | .801 | .941 | .865 | .795 | .933 | .858 |
| .995 | .818 | .928 | .869 | .812 | .922 | .863 |
| .999 | .847 | .919 | .881 | .843 | .902 | .871 |

Table 1: Comparison of results: Multiple Corrections per Window Dashti (2017), (Wilcox-O'Hearn and Hirst (2008), Hirst and Budanitsky (2005) (shown on the first row) and the proposed approach on Wall Street Journal corpus with a 62,000 word vocabulary

The reason for this observation is that in situations where α has a lower value, the probability that a current window incorporates at least two real-word errors would notably increase. However, what distinguishes our approach from Dashti's (2017) model is the kind of syntactic knowledge we use. Initially the proposed model can detect almost all morphological analyses (over 99%) of a token in a correction candidate. Next, the method manipulates thousands of handwritten grammar rules and heuristics to give unambiguous analyses of tokens with 99.7% accuracy, if the correct analysis exists. A good correction candidate is supposed to be syntactically correct, if all the tokens accommodated in it are unambiguous. For higher values of α (e.g. α=.995), the models of Wilcox-O'Hearn and Hirst (2008) and Dashti (2017) showed nearly similar results. The proposed model, however, still performed significantly better. This performance level, of course, was not very unexpected, because the syntactic knowledge proved to be practical even in correcting single real-word errors in a window. Yet, the results of the proposed approach on the MALP test set were still good in comparison to the S62000 test set. This good performance, too, was also anticipated. Although the MALP test set included content-word errors, CG helped to detect these types of syntactically ill-formed errors. Nevertheless, the results of the proposed method showed noticeably much better performance in contrast with those of Hirst and Budanitsky's (2005) model (see the first row of Table 1). It should be noted that for $d=1$ span of words, where window-size=5, the search space $C(S)$ incorporated an average of 29 sentences. Then, $d$ was expanded to higher values of 3, 6 and 10. However`, as expected the results were only slightly changed; since all possible correction candidates were generated in the same way and then analyzed by using CG. Table 2 shows the results for d=3, d=6 and d=10.

|  | Detection | | | Correction | | |
|---|---|---|---|---|---|---|
| α | P | R | F | P | R | F |
| Test set **S62000,** *d*=3: | | | | | | |
| .9 | .777 | .966 | .861 | .770 | .948 | .850 |
| .99 | .804 | .950 | .870 | .797 | .941 | .863 |
| .995 | .818 | .939 | .874 | .811 | .933 | .867 |
| .999 | .847 | .923 | .883 | .844 | .909 | .875 |
| Test set **MALP,** *d*=3: | | | | | | |
| .9 | .773 | .957 | .855 | .767 | .937 | .843 |
| .99 | .802 | .939 | .865 | .794 | .931 | .857 |
| .995 | .816 | .929 | .868 | .810 | .921 | .861 |
| .999 | .848 | .917 | .881 | .842 | .900 | .870 |
| Test set **S62000,** *d*=6: | | | | | | |
| .9 | .774 | .967 | .859 | .769 | .949 | .849 |
| .99 | .805 | .948 | .870 | .798 | .939 | .862 |
| .995 | .816 | .940 | .873 | .809 | .935 | .867 |
| .999 | .848 | .921 | .882 | .846 | .911 | .877 |
| Test set **MALP,** *d*=6: | | | | | | |
| .9 | .776 | .955 | .856 | .773 | .938 | .847 |
| .99 | .802 | .942 | .866 | .791 | .935 | .856 |
| .995 | .820 | .931 | .871 | .813 | .929 | .867 |
| .999 | .849 | .920 | .883 | .841 | .904 | .871 |
| Test set **S62000,** *d*=10: | | | | | | |
| .9 | .771 | .968 | .859 | .767 | .948 | .847 |
| .99 | .803 | .945 | .868 | .796 | .937 | .860 |
| .995 | .815 | .939 | .872 | .807 | .933 | .865 |
| .999 | .850 | .921 | .884 | .844 | .901 | .871 |
| Test set **MALP,** *d*=10: | | | | | | |
| .9 | .780 | .960 | .860 | .772 | .935 | .845 |
| .99 | .809 | .943 | .870 | .799 | .933 | .860 |
| .995 | .819 | .934 | .869 | .812 | .926 | .865 |
| .999 | .852 | .914 | .881 | .846 | .898 | .871 |

Table 2: Evaluating the Hybrid approach by using values of *d*=3 *d*=6 *d*=10, on **S62000** and **MALP** test set.

Apparently, as the values of *d* increased, the measure of recall decreased, but in the meantime precision significantly rose up. In Table 3 some examples of multiple successful and unsuccessful corrections in a particular window are provided.

| Successful multiple corrections |
|---|
| … can almost see the *firm* → farm [farm] issue *seceding* → receding [receding]… |
| … again are confronting a *bell* → ball [ball] game in which they will be able *too* → to [to] play… |
| they came on the heels of the Reykjavik *summits* → summit [summit] between President Reagan and Soviet *leaders* → leader [leader] Mikhail Gorbachev. |
| Unsuccessful multiple corrections |
| True Positive correction of one error; False Negative detection one error: … I'm uncomfortable *tacking* → taking [taking] a lot of *times* [time] off work," he says. |

Table 3: Examples of successful and unsuccessful multiple corrections. Italics demonstrate the words which are thought to be errors, arrow demonstrate the correction replaced by the error, and string inside brackets show the intended word.

The evaluations proved that the proposed model was successful in accomplishing the task of correcting multiple real-word errors in a particular window. Meanwhile, during the assessments, as the span of words, *d* was expanded and the windows accommodated an increased number of words, the runtime increased and the model showed some performance overhead. The reason for this was that the model had to deal with numerous correction candidates and to apply CG on each, to discriminate among them. Hardware platform used in this comparison was exactly the same as the one employed by Dashti (2017): HPE ProLiant ML150 Gen9 Server model; with Intel Xeon E5-2600 v4 Processor and 256 GB RAM (DDR4- 21,400 MHz).

Table 4 demonstrates a comparison of average correction time of all the test instances, for different values of the parameter *d* between the proposed model and those of Wilcox-O'Hearn and Hirst (2008) and Dashti (2017).

| d | Correction time (in milliseconds) | | |
|---|---|---|---|
| | Hybrid Approach | Fixed window size | Multiple corrections per window |
| 1 | 910 | 581 | 690 |
| 3 | 1026 | 781 | 871 |
| 6 | 1133 | 991 | 1357 |
| 10 | 1409 | 1230 | 2014 |

Table 4: Average correction time of all the test sentences for different values of parameter *d*

Although for lower values of the parameter *d*, more iterations would be required to cover word- sequences in a sentence, the runtime would be considerably better because fewer combinations would be generated in each window.

Complete information regarding the size of the search space is presented here. Table 5 represents the average size of the search space according to the window size. Unlike Dashti (2017) model, no significant difference might be seen, because all possible correction candidates were generated either by using smaller windows or larger windows as described in 3.4 section. After that, the CG was applied to all correction candidates which were accommodated in the initial search space. This process ultimate led to the final search space, which included only syntactically correct candidates. The good performance of the model revealed that the model could be completely practical and efficient in everyday applications for all types of users.

| Parameter *d* | Initial Search Space Size | Final Search Space Size |
|---|---|---|
| 1 | 5230 | 29 |
| 3 | 5078 | 33 |
| 6 | 4920 | 27 |
| 10 | 5001 | 30 |

Table 5: Average size of the search space

# 6 Other relevant observations

Introducing the notion of *malapropism*, Hirst and Budanitsky (2005) tried to find a way to recognize and correct any anomalous words in a text. This method rested on the lexical-resource of WordNet. They manipulated the measure of lexical cohesion to identify the semantic distance in the text. In cases where spelling variation led to a word that displayed a semantic content matching the context, the method would assume that the original word was an error. Furthermore, Hirst and Budanitsky's model failed to perform better than other alternatives, such as models developed by Mays et al. (1991), Wilcox-O'Hearn and Hirst (2008), and Dashti (2017). Some studies have also dealt with correcting real-word errors (see Golding and Schabes, 1996; Golding and Roth, 1996; Golding and Roth, 1999). These investigations relied on machine learning techniques, processing real-word error correction as a function of disambiguation. Another model to trace anomalies between words involves *predefined confusion sets* in a particular lingual context (Golding and Roth 1996). These sets were extracted from the list of usually confused words, as enumerated by Random House Unabridged Dictionary Flexner (1983). What distinguished these three methods (Golding and Schabes, 1996; Golding and Roth, 1996; Golding and Roth, 1999) was the specific techniques employed to deal with the real-word error correction task. Golding and Roth (1996, 1999), manipulated WinSpell software, which drew on a machine-learning algorithm. In this configuration, the members of confusion sets were represented as clouds of "slow neuron-like" nodes, which reflected collocational groupings and repeated features. Furthermore, Golding and Schabes (1996) incorporated a Bayesian hybrid strategy, and Golding (1995) employed a pos-trigram model. Newer models were suggested by Fossati and Di Eugenio (2007, 2008). These researchers configured a mixed trigram method that utilized the data of a part of speech tagger. Mixed trigrams were included grammatical categories as well as words (e.g. articles, adjectives, verbs). As a result of this process, fewer trigrams were needed to be generated. According to mixed trigrams' indications, each word in a sentence was parceled to reveal its correct grammatical order. The analyzed word could be associated with some candidates (from a confusion set) that had to be verified. Through mixed trigrams, a word as suggested by the confusion set was selected as the proper alternative. Verberne (2002) also suggested another method: as her hypothesis postulated, any word-trigram in the context of the British National Corpus (BNC) was correct, and any trigram that was not found in BNC would be undoubtedly an error. In cases where an unspecified word-trigram was to be verified, the model tried to go through the possible spelling variations of the current words in the trigram to detect relevant trigrams. She obtained a recall of .33 and a precision of .05.

## 7 Conclusion

This study proposed a method which could provide a significantly better performance compared with the trigram approach of Mays et al. (1991), windows of fixed size model of Wilcox-O'Hearn and Hirst (2008), and multiple correction approach proposed by Dashti (2017). The study demonstrated that the proposed model yielded significantly better accuracy in correcting multiple errors in a current window for lower values of α. In contexts with higher values of α, the difference was still noticeable and the proposed approach outperformed the alternative models. In terms of running time, the average error correction time increased for larger values of the parameter $d$ and the performance was noticeably lower. In case of correcting malapropisms, as observed by Mays et al. (1991), Wilcox-O'Hearn and Hirst (2008), and Dashti (2017), the proposed model demonstrated an exceptionally good performance, particularly where the number of errors was considerably higher (lower values of α). Our attempt to improve the real-word error correction task, through statistical and syntactic knowledge, was completely successful.